# Acquiring Common Chinese Emotional Events Using Large Language Model

YA WANG, School of Artificial Intelligence and Software Engineering, Nanyang Normal University, Nanyang, China, wangya@nynu.edu.cn

GUANGZHENG ZHU, School of Artificial Intelligence and Software Engineering, Nanyang Normal University, Nanyang, China, zhuguangzheng@nynu.edu.cn

CUNGEN CAO, Key Laboratory of Intelligent Information Processing, Institute of Computer Technology, Chinese Academy of Sciences, Beijing, China, cgcao@ict.ac.cn

JINGJING LI, School of Artificial Intelligence and Software Engineering, Nanyang Normal University, Nanyang, China, jingjl101@nynu.edu.cn

HE LI, School of Artificial Intelligence and Software Engineering, Nanyang Normal University, Nanyang, China, lihe@nynu.edu.cn

XIN HUANG* School of Artificial Intelligence and Software Engineering, Nanyang Normal University, Nanyang, China, huangxin@nynu.edu.cn

Knowledge about emotional events is an important kind of knowledge which has been applied to improve the effectiveness of different applications. However, emotional events cannot be easily acquired, especially common or generalized emotional events that are context-independent. The goal of this paper is to obtain common emotional events in Chinese language such as "win a prize" and "be criticized". Our approach begins by collecting a comprehensive list of Chinese emotional event indicators. Then, we generate emotional events by prompting a Chinese large language model (LLM) using these indicators. To ensure the quality of these emotional events, we train a filter to discard invalid generated results. We also classify these emotional events as being positive events and negative events using different techniques. Finally, we harvest a total of 102,218 high-quality common emotional events with sentiment polarity labels, which is the only large-scale commonsense knowledge base of emotional events in Chinese language. Intrinsic evaluation results show that the proposed method in this paper can be effectively used to acquire common Chinese emotional events. An extrinsic use case also demonstrates the strong potential of common emotional events in the field of emotion cause extraction (ECE). Related resources including emotional event indicators and emotional events will be released after the publication of this paper.

CCS CONCEPTS: • Computing methodologies →Artificial intelligence →Natural language processing; Language resources.

Additional Keywords and Phrases: common Chinese emotional event, emotional event indicator, emotional event acquisition, emotion cause extraction.

---

Corresponding Author

# 1 INTRODUCTION

Many events occurring in our daily life affect us in positive or negative ways. For example, we will feel happy when we win an international prize and become sad when we are criticized by someone. We refer to an event that provokes a particular emotion in the person who experiences it as emotional event or emotion-provoking event [1]. Recognizing emotional events and their sentiment polarity is essential for many natural language processing tasks, including response generation [2], implicit sentiment analysis [3], emotion cause extraction [4] and other applications related to sentiment analysis. However, there is a lack of common Chinese emotional events, which hinders related research in Chinese language.

Through a linguistic analysis of Chinese emotional events, we discover that many Chinese emotional events are expressed using emotional event indicators such as "被 (*bei*)", "获得 (*huo de*)" and "丢失 (lose)". These indicators are linguistic cues that mark the presence of emotional events. For simplicity, we will refer to emotional event indicators as indicators in the rest of this paper. We can divide Chinese emotional events into explicit emotional events and implicit emotional events based on these indicators. Explicit emotional events are composed of indicators and the corresponding themes (e.g. "被批评 (be criticized)", "获得奥运金牌 (win an Olympic gold medal)" and "丢失一个手机 (lose a cell phone)"), and implicit emotional events are emotional events without any indicators (e.g. "发财 (make a fortune)" and "失败 (fail)"). The number of explicit emotional events is more than that of implicit emotional events. The focus of our work is on explicit emotional event acquisition. It is worth pointing out that, each Chinese character or word is paired with the corresponding *Chinese pinyin* in italics or English translation for its first occurrence in this paper.

With the recent surge in the development of generative large language models (LLMs), a large body of literature has emerged that employs LLMs to generate commonsense knowledge [5][6]. Motivated by the characteristics of Chinese emotional events and the remarkable generation ability of LLMs, we propose a novel method to generate common Chinese emotional events by prompting an off-the-shelf instruction-following LLM. We use the LLM in a few-shot setup, where only a small amount of labeled data (examples) is required compared to fine-tuning it. Through pilot studies, we empirically choose SparkDesk (i.e. 讯飞星火 in Chinese) [7] for emotional event generation. SparkDesk is one of the popular and biggest publicly available Chinese LLMs. To our knowledge, there is no prior work on acquiring common emotional events using LLMs. Our method starts with collecting a comprehensive list of Chinese emotional event indicators, which is a key challenge of our work. We harvest a total of 726 indicators through manual extraction and construction. We further categorize these indicators into three classes: positive indicators, negative indicators and neutral indicators. For example, the aforementioned "获得" is a positive indicator, "丢失" is a negative indicator and "被" is a neutral indicator. We then exploit SparkDesk to generate common emotional events by designing natural language prompts based on these indicators. We provide a corresponding prompt for each indicator, with which we can obtain 100 events. Among these indicators, "被" is a special indicator for its neutral polarity. We call an emotional event involving "被" as *bei* event in the rest of this paper. In our daily life, the number of *bei* events is much more than 100. In order to acquire more *bei* events, we combine "被" and an appropriate verb into a compositional indicator like "被没收 (be confiscated)" and "被禁止 (be forbidden)". These compositional indicators involving "被" are exclusive indicators of *bei* events. For this reason, we refer to these indicators of *bei* events as *bei* event indicators. In this way, we can increase the number of indicators containing "被" and thus the number of prompts. Consequently, we are allowed to acquire more *bei* events with more prompts. In order to further improve the quality of these LLM-generated emotional events, we develop a binary classifier to filter the incorrect emotional events. Finally, we acquire 102,218 high-quality common Chinese emotional events, which is the first Chinese resource of common emotional events.

We conduct a human evaluation on the quality of these emotional events. Intrinsic evaluation results show that our method is effective for acquiring common Chinese emotional events with a high precision of 0.96. The ECE or emotion cause detection task aims to identify clauses which contain emotion-provoking information for a particular emotion expressed in text [8][9][10]. Furthermore, we evaluate ConceptNet [11], the most prominent commonsense knowledge graph, on its coverage of emotional event nodes. Unfortunately, ConceptNet has an extremely low emotional event coverage of 0.01, which validates that existing commonsense knowledge graphs are limited to emotional events and our acquired common emotional events can be used to enrich existing commonsense knowledge graphs by adding novel emotional event nodes. To investigate the application of our emotional events about their effectiveness for ECE, we enhance eight representative emotion cause extraction models with them. Then, we evaluate the performance of these models and their corresponding augmented models on a Chinese emotion cause corpus.

In conclusion, the main contributions of our work are as follows: 1) the first to use a LLM to automatically generate common emotional events in a few-shot setup, 2) provides a comprehensive list of Chinese emotional event indicators for further emotional event acquisition, 3) constructs a large-scale and high-quality knowledge base of common emotional events for Chinese applications, especially Chinese sentiment analysis, and 4) conducts comprehensive experiments to demonstrate the effectiveness of the collected emotional events in the field of emotion cause extraction.

The rest of this paper is organized as follows. Section 2 reviews the related work on emotional event acquisition and emotion cause extraction. In section 3, we present the methods of acquiring and categorizing common Chinese emotional events. Intrinsic evaluation results and extrinsic evaluation results are reported and discussed in section 4. Finally, section 5 summarizes our work about its limitations and future research directions.

## 2 RELATED WORK

In this section, we will review prior work on emotional event acquisition and emotion cause extraction, which are closely related to our work in this paper.

### 2.1 Common emotional event acquisition

To acquire knowledge about stereotypically positive and negative events, [12] explored semi-supervised learning with graphs using a large collection of personal blogs. A sentiment classifier was applied to identify sentences that have strong positive or negative polarity. [13] introduced a weakly supervised method to infer affective events from a personal story corpus. [14] proposed an attention-based neural network model to identify implicit polarity of events. In contrast to [12], [13] and [14], our emotional event acquisition method is fully unsupervised that requires no human-annotated samples. In order to extract major life events, [15] proposed a pipelined system. The system can automatically identify individual life events, e.g. receive award and get married. However, the system achieves a much lower precision (0.64) than our method. With the purpose of promoting a more holistic approach to sentiment analysis, the SemEval-2015 Task 9 [16] provided a dataset of events manually annotated as instantiations of pleasant and unpleasant events. The size of the final dataset is 1,651 events, which is much smaller than our dataset containing more than 100,000 events. [3] constructed a dataset called EveSA for implicit sentence-level sentiment analysis. In EveSA, each sentence is associated with its sentiment polarity (i.e. positive, negative and neutral). The 18 event types that have obvious sentiment inclinations are manually defined based on FrameNet [17], an English vocabulary knowledge base. Our emotional events cover much more event types compared to events in this paper. [10] introduced connotation frames as a representation framework to encode the rich dimensions of the implied sentiment and presupposed facts for a verb predicate. Given a predicate, four different typed relations are defined, including perspective, effect, value and mental state. In order to induce the connotation frames of verb predicates, [18] proposed an aspect-level model based on the distributional word representations and a frame-level model using the interplay between different typed relations. In contrast to our events with richer structures, [18] focused on individual verbs. [19] recognized emotional events based on lexico-syntactic patterns associated with first-person affect. A weakly supervised pattern learner was employed to produce new lexico-syntactic patterns to expand the seed first-person sentiment corpus. Since these patterns are automatically learned, they are noisy. Also, the number of these patterns is limited. [20] presented a corpus consisting of <cause, emotion, action> triplets, e.g. <listening to what I said, happy, joked with me>. In a triplet, the cause refers to the event that evokes the emotion and the action represents the event that is triggered by the emotion. The cause in a triplet is an emotional event. However, most of these emotional events are not common emotional events. In addition, these emotional events are annotated manually. [21] prompted pretrained language models to elicit new affective events. However, most of the generated events are not affective events (e.g. "I go to hotel" or "eat my lunch").

Although there are a number of studies on acquiring English emotional events, we are not aware of existing research on emotional event acquisition in Chinese language except for [20]. In terms of common emotional events, as far as we know, this paper is the first to propose to acquire common Chinese emotional events.

## 2.2 Emotion cause extraction

Earlier studies of emotion cause extraction mainly employed rule-based methods [22] [23]. [23] introduced a multi-label approach to recognize emotion causes. The multi-label model can detect emotion causes and capture the long-distance information to facilitate emotion cause detection as well. To extract emotion cause, [22] constructed a Chinese emotion cause corpus and designed two sets of linguistic rules for emotion cause extraction based on this corpus. However, these rule-based emotion cause extraction methods suffer from poor extraction performance because the human-created rules are limited in number and cannot cover all complex linguistic phenomena of texts in practical applications. To overcome the limitations of rule-base methods, a variety of machine learning methods were proposed for ECE task. [24] proposed a conditional random field (CRF) model based on the syntactic and semantic characteristics of Weibo texts. This model can effectively mine the relation between an emotion and its causes in Weibo texts. [25] presented an Emotion Cause-OCC model to identify emotion causes in Chinese microblog posts. This model focused on mining factors eliciting particular kinds of emotions and could acquire the proportions of these cause components under different emotions. [26] addressed the emotion cause extraction problem as information extraction and built a CRF learner to detect the emotion stimuli spanning in emotion-bearing sentences. [4] proposed an event-driven emotion cause extraction method using multi-kernel SVMs. In this method, a syntactical tree-based approach is exploited to represent events in text. With the development of deep learning techniques, many deep learning methods for emotion cause extraction have been proposed. [27] proposed a co-attention deep neural network for emotion cause recognition by exploiting the correlation between clauses. Following this work, [28] proposed a multi-attention-based neural network (MANN) to mine correlations between emotion phrases and candidate clauses. Subsequently, [29] presented a model, called RHNN (regularized hierarchical neural network), to extract emotion cause by incorporating discourse context information and using a sentiment lexicon and commonsense knowledge as restrained parameters. [30] presented a joint emotion cause extraction framework, named RTHN (RNN-Transformer Hierarchical Network). RTHN consists of a lower word-level encoder based on RNNs and an upper clause-level encoder based on Transformer. [31] considered emotion cause detection task as a reading comprehension task in question-answering and built a QA system based on a deep memory network. Both machine learning methods and deep learning methods require human-annotated training data which is expensive to obtain.

Previous studies on emotion cause extraction use either unsupervised rule-based methods or supervised machine learning and deep learning methods. However, they ignore the valuable resource of emotional events especially common emotional events which are important emotion causes themselves and highly useful in emotion cause recognition.

## 3 COMMON EMOTIONAL EVENT ACQUISITION AND CLASSIFICATION

In this section, we will describe the methodology of common Chinese emotional event acquisition and classification in detail. The overall process is shown in Figure 1.

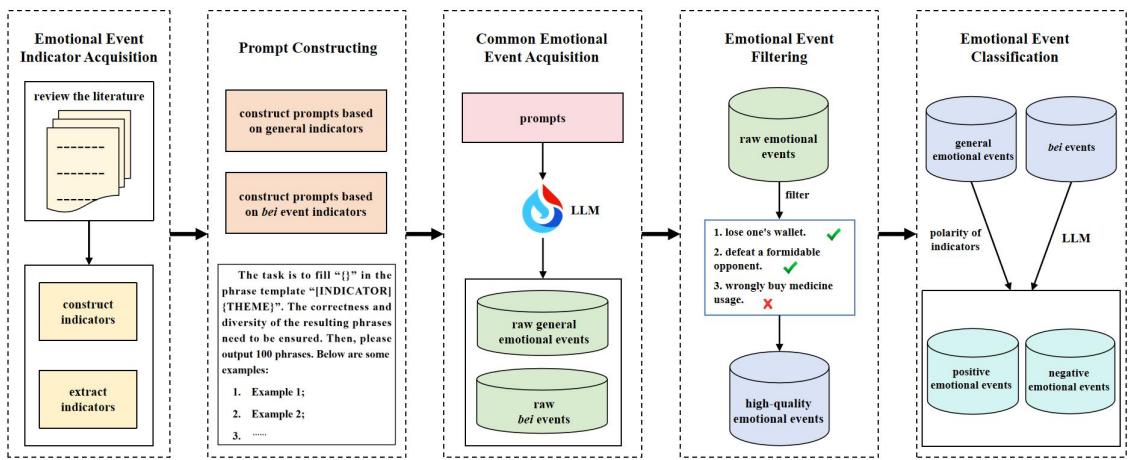

Figure 1: Flowchart of the overall method.

### 3.1 Acquisition and classification of emotional event indicators

A key challenge of our work is to obtain emotional event indicators. Some Chinese emotional events have been analyzed in the previous literature [32][33][34]. We perform a thorough review of the literature and extract indicators from the literature or construct the indicators we need based on the literature. For example, we extract a sublist of indicators represented in the form of "错 V (e.g. 错爱 (love the wrong person), 错拿 (take the wrong item) and 错吃 (eat the wrong food))" based on [23] and construct a sublist of indicators expressed in the form of "漏 V (e.g. 漏查 (omit to check), 漏标 (omit to mark) and 漏关 (omit to close))" based on [35], where V denotes a verb. This step contributes 430 indicators. In addition, we find that most Chinese resultative verbs such as "踩坏 (step on and damage)" and "压坏 (press and damage)" are composed of a verb representing an action and the outcome caused by this action. The outcome or result in a resultative verb is usually out of the control of human beings, especially the bad one. Therefore, we argue that Chinese resultative verbs can be considered as emotional event indicators. [36] is a book that contains 322 verbs, adjectives and phrases that frequently occur as resultative complements. We create appropriate verb-resultative complement phrases or resultative verbs as emotional event indicators based on this book. This creation step produces a sublist of 296 indicators. For neutral indicators, we only add "被" to our indicator list, as it is the most frequently used neutral indicator. Finally, we acquire a comprehensive list of 726 Chinese emotional event indicators.

In the process of acquiring emotional event indicators, we discover that some indicators have weak sentiment intensity (e.g. "穿皱 (get wrinkled from wearing)" and "骑脏 (get dirty from riding)") and we refer to these indicators as weak indicators. Accordingly, we regard emotional events with weak indicators as weak emotional events, e.g. "穿皱衣

服 (wear one's clothes wrinkled)" and "骑脏自行车 (ride one's bike and then get the bike dirty)". People usually have no emotional reactions when these weak emotional events happen to them. In other words, weak emotional events contribute little to people's emotional state and thus to related sentiment analysis tasks neither. For this reason, we remove these weak indicators from our indicator list. Also, some indicators are ambiguous. For example, for the event "白听一场演唱会 (*bai ting yi chang yan chang hui*)" with indicator "白听", we can comprehend it as a person listen to a concert paying no money (false emotional event) or a person listen to a concert learning nothing (true emotional event). To reduce LLM hallucinations when we use SparkDesk to generate common emotional events in the next section, we also delete these indicators from the indicator list.

Table 1: Statistics of Chinese emotional event indicators.

| Indicator Types | Examples of Indicators | Number of Indicators |
|---|---|---|
| classic indicators | 遭到 (*zao dao*), 遭受 (*zao shou*), 忍受 (*ren shou*), 饱受 *(bao shou)*, 蒙受 (*meng shou*) | 7 |
| neutral indicators | 被 | 1 |
| resultative verbs | 踩坏, 炒糊 (overcook and scorch), 考砸, 凿偏 (strike askew), 烤坏 (overgrill and ruin) | 296 |
| "白 V" | 白写 (write in vain), 白等 (wait in vain) 白去 (go in vain), 白干, 白说 (say in vain), | 13 |
| "V 破" | 穿破 (wear through), 洗破 (wash and tear), 挤破 (squeeze and burst), 撑破 (stretch and rip), 磕破 (bump and break) | 41 |
| "错 V" | 错喊 (miscall), 错拿, 错吃, 错嫁 (marry the wrong person), 错报 (misreport) | 30 |
| "V 错" | 喊错 (miscall), 拿错 (take the wrong item), 吃错 (eat the wrong food), 嫁错 (marry the wrong person), 报错 (misreport) | 99 |
| "V 对" | 猜对 (guess correctly), 算对 (calculate correctly), 改对 (make the right correction), 写对 (write correctly), 读对 (read correctly) | 16 |
| "漏 V" | 漏拔 (omit to pull), 漏查, 漏关, 漏标, 漏测 (omit to measure) | 136 |
| others | 摆脱 (shake off), 获得, 荣获 (be honored with), 痛失 (lose), 免于 (be exempt from) | 87 |

Table 1 presents the statistics of all indicators we obtain. The majority of indicators have to co-occur with their topics or themes to produce semantically complete emotional events. Some indicators themselves are abstract emotional events, e.g. "白干 (work in vain)" and "考砸 (fail the exam)". It is noteworthy that indicators in the forms of "V 破", "V 错" and "V 错" are not contained in resultative verbs.

Table 2: Classification of indicators.

| Positive Indicators | Neutral Indicators | Negative Indicators |
|---|---|---|
| 摆脱 | 被 | 遭受 |
| 获得 | -- | 面临 (be confronted with) |
| 荣获 | -- | 丢失 |
| 打败 (defeat) | -- | 错吃 |

| Positive Indicators | Neutral Indicators | Negative Indicators |
|---|---|---|
| 免于 | -- | 白等 |
| 142 | 1 | 583 |

After indicator acquisition, we categorize these indicators into three classes (positive, negative or neutral). Since the size of the indicator list is small, we decide to manually classify these indicators to achieve the best classification performance. We ask the first author of this paper who is familiar with these indicators to carry out this task. We decompose the indicator classification task into two subtasks: (1) Firstly, categorizing the indicators having the same lexical structure, namely "白 V", "错 V", "V 错" and "漏 V"; (2) Secondly, dealing with the remaining indicators. Both subtasks can be completed in less than an hour, especially the first one. The affective polarity distribution of these indicators is reported in Table 2.

### 3.2 Emotional event acquisition using LLM

As mentioned earlier, LLMs have powerful text generation abilities and can generate coherent and meaningful outputs based on prompts or contextual inputs. One of the key challenges in utilizing LLMs lies in how to effectively interact with them. This primarily involves designing appropriate prompts that can guide the models to produce desired outputs. Since the outputs of an LLM is generated in response to the input prompts, prompt designing becomes a critical step that influences the quality, relevance and task alignment of the generated content. A well-crafted prompt can lead to accurate and informative responses, whereas a poorly designed prompt may result in outputs that deviate from expectations or contain vague and ambiguous content. Consequently, constructing effective prompts has become a central issue in the practical application of LLMs. Researchers have proposed a variety of prompting strategies, including zero-shot prompting, few-shot prompting, knowledge-enhanced prompting, chain-of-thought prompting and tree-of-thought prompting. Few-shot prompting is characterized by using selected examples and combining with a natural language description of the task to directly guide the models to generate new responses, without requiring additional fine-tuning.

> The task is to fill "{}" in the phrase template "[INDICATOR] {THEME}". The correctness and diversity of the resulting phrases need to be ensured. Then, please output 100 phrases. Below are some examples:
>
> 1. Example 1;
> 2. Example 2;
> 3. Example 3;
> 4. Example 4;
> 5. Example 5;
> 6. Example 6;
> 7. Example 7;
> 8. Example 8.

Figure 2: Prompt template.

In this paper, we adopt a few-shot prompting strategy to generate common emotional events, wherein a small number of human-crafted examples are provided to assist the models in performing a specific task. These examples serve to reduce ambiguity and enhance the model's understanding of the task. Providing examples is an effective means of prompt optimization. In more detail, we construct corresponding prompts based on the indicators we obtain in section 3.1 and use these prompts to elicit common emotional events from an off-the-shelf LLM (in our case, SparkDesk). The prompt template is presented in Figure 2, where [INDICATOR] is a placeholder that can be replaced with one of the 726 indicators and {THEME} is a placeholder to be filled with possible themes SparkDesk produces given an indicator. Concrete prompts will be given in the following sections. We conduct a preliminary investigation to examine the ability of various LLMs to generate common emotional events with these prompts. We compare the generation performance of five Chinese LLMs (i.e. DeepSeek, SparkDesk, ERNIE Bot (i.e. 文心一言 in Chinese), ChatGLM (i.e. 智谱清言 in Chinese) and Kimi) and two English LLMs (i.e. GPT3.5-turbo and GPT-4o), from which we observe that SparkDesk produces less hallucinated events than other LLMs. Therefore, we select SparkDesk to acquire emotional events. We first acquire emotional events with positive or negative indicators in section 3.2.1. Then, we collect emotional events with *bei* event indicators in section 3.2.2.

---

The task is to fill "{}" in the phrase template "遭受{}". The correctness and diversity of the resulting phrases need to be ensured. Then, please output 100 phrases. Below are some examples:

1. 遭受挫折 (suffer a setback);
2. 遭受好朋友的背叛 (betrayed by a good friend);
3. 遭受陌生人的袭击 (attacked by strangers);
4. 遭受校园暴力 (suffer from school violence);
5. 遭受无端的网络暴力 (subjected to cyber violence for no reason);
6. 遭受社会的不公正对待 (suffer from unfair treatment by society);
7. 遭受灭顶之灾 (suffer a catastrophe);
8. 遭受失眠的困扰 (suffer from insomnia).

---

Figure 3: Prompt for indicator "遭受".

Table 3: Examples of emotional events with indicator "获得".

| | | |
|---|---|---|
| 获得全额奖学金<br>(receive a full scholarship) | 获得朋友的支持<br>(get support from friends) | 获得科研经费<br>(get research funding) |
| 获得荣誉称号<br>(win an honorary title) | 获得国际大奖<br>(win an international prize) | 获得博士学位<br>(get one's PhD) |
| 获得更优厚的待遇<br>(get a better treatment) | 获得家人的理解<br>(own understanding from the family) | 获得客户的信任<br>(win the trust of customers) |

### 3.2.1 Acquire emotional events involving positive or negative indicators.

This section describes the procedure of acquiring emotional events containing positive or negative indicators using SparkDesk. We find in our preliminary experiments that, when the number of events generated by SparkDesk exceeds 100, incorrect generated events begin to appear for most indicators. For this reason, we set the maximum number of the generated events to 100 to ensure the precision of these events. Moreover, [37] and [38] show that better examples given in the prompts will lead to a performance boost for incontext learning. Therefore, we provide 8 few-shot human-curated examples for each prompt to ensure the quality of these examples. These examples are used to help SparkDesk to better comprehend and respond to the prompts. We iterate over all non-neutral indicators to construct specific prompts and feed these prompts to SparkDesk to obtain new emotional events. An example prompt is illustrated in Figure 3.

However, due to the hallucination problem in LLMs, SparkDesk sometimes produces duplicate events or generates fewer than 100 events for some indicators. Finally, we collect a total of 52,512 emotional events for 725 non-neutral indicators by removing the duplicate events. 9 of the events with indicator "获得" are given in Table 3.

### 3.2.2 Acquire emotional events involving bei event indicators.

For the neutral indicator "被", we can only acquire 100 events using the prompt, which is far less than the number of *bei* events. In order to acquire more *bei* events, we combine "被" and a verb as an indicator (i.e. *bei* event indicator), e.g. "被喜欢 (be liked)" and "被淘汰 (be eliminated)". By doing so, we can increase the number of indicators involving "被" and thus the number of corresponding prompts. We are allowed to acquire more *bei* events with more prompts. However, the most important and challenging task is to obtain *bei* event indicators, namely the collocations of "被" and its followed verbs. It is worth mentioning that, not all Chinese verbs can be collocated with "被" to make a correct *bei* event indicator. For example, "被批评 (be criticized)" is a correct indicator, but "被上当 (*bei shang dang*)" is an incorrect indicator. Since the number of Chinese verbs is limited, we decide to manually construct the combinations of "被" and the verbs to ensure the quality of them. Finally, we select 918 verbs that can be combined with "被" based on [35] and increase the number of indicators containing "被" from 1 to 918. Examples of *bei* event indicators and a corresponding prompt are presented in Table 4 and Figure 4, respectively.

Table 4: Examples of *bei* event indicators.

| 被禁止<br>(be banned) | 被控制<br>(be controlled) | 被忽悠<br>(be deceived) | 被开除<br>(be dismissed) |
|---|---|---|---|
| 被救助<br>(be rescued) | 被攻击<br>(be attacked) | 被列入<br>(be included in) | 被没收 |
| 被拒绝<br>(be rejected) | 被报复<br>(be retaliated against) | 被扣除<br>(be deducted) | 被善待<br>(be treated kindly) |

<table>
<tr><td>

The task is to fill "{}" in the phrase template "被禁止{}". The correctness and diversity of the resulting phrases need to be ensured. Then, please output 100 phrases. Below are some examples:

1. 被禁止参加国际比赛 (be banned from international competitions);

2. 被禁止乘公共汽车 (be forbidden to ride the bus);

3. 被禁止营业 (be banned from doing business);

4. 被禁止进入考场 (be barred from entering the examination room);

5. 被禁止留长头发 (be forbidden to have long hair);

6. 被禁止发言 (be forbidden to speak);

7. 被禁止使用手机 (be forbidden to use cell phones);

8. 被禁止进入图书馆 (be barred from entering the library).

</td></tr>
</table>

Figure 4: Prompt for bei event indicator "被禁止".

Similar to acquiring emotional events with non-neutral indicators, acquiring *bei* events also suffers from the hallucination issue in LLMs, i.e. SparkDesk outputs duplicate events or generates fewer than 100 events. Since "被" itself is a neutral indicator, the polarity of some *bei* event indicators is neutral as well, such as "被安排 (be arranged)" and "被模仿 (be imitated)". Consequently, the generated events with such *bei* event indicators can be neutral events, e.g. "被安排参加会议 (be arranged to attend a meeting)" and "被模仿舞蹈动作 (be imitated in dance moves)". The aim of this paper is to acquire emotional events whose polarity is non-neutral. For this purpose, we need to discard these neutral *bei* events. We perform this task by asking SparkDesk if the polarity of a generated *bei* event is neutral. If so, we discard it. Finally, we obtain a total of 68,488 *bei* events and 9 of the events with indicator "被没收" are described in Table 5.

Table 5: Examples of *bei* events with indicator "被没收".

| 被没收财产<br>(have one's property<br>confiscated) | 被没收手机<br>(get one's mobile phone<br>confiscated) | 被没收护照<br>(have one's passport<br>confiscated) |
|---|---|---|
| 被没收钥匙<br>(get one's key confiscated) | 被没收信用卡<br>(have one's credit card<br>confiscated) | 被没收电子设备<br>(get one's electronics<br>confiscated) |
| 被没收电脑<br>(have one's computer<br>confiscated) | 被没收首饰<br>(have one's jewelry<br>confiscated) | 被没收书籍<br>(get one's books confiscated) |

In the process of constructing *bei* event indicators, we also identify 182 classic implicit emotional events which contain no indicators, like "获奖 (win a prize)" and "摔倒 (fall down)". Examples of these implicit emotional events are given in Table 8. Finally, we acquire 121,182 raw common Chinese emotional events in total.

### 3.3 Emotional event filtering

As stated before, LLMs can generate false statements about common emotional events due to the hallucination issue. To address this, we train a binary classifier (filter) to predict the validity of each generated statement.

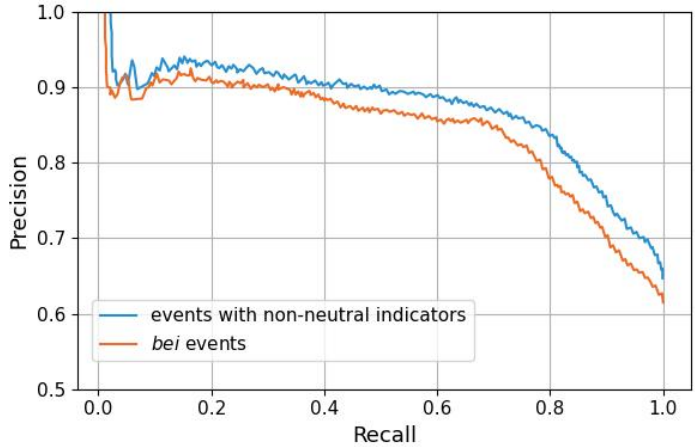

Figure 5: The precision-recall curves of our filter model on the human labelled validation set.

We randomly sample 10 instances for each indicator from the raw generated emotional events and ask three native Chinese speakers to judge whether an emotional event is valid. The annotators are Chinese university students majoring in Chinese Language and Literature who are highly familiar with Chinese emotional events. If an emotional event is linguistically and semantically correct, we assign a "valid" to it; Otherwise, we label it with "unvalid".

We use 80% of the annotated data for training, and the remaining data for validating and testing. We train a RoBERTa-Large classifier [39] as our filter to identify valid emotional events. We conduct a small grid search on the validation set and find batch size 128, dropout 0.1 and learning rate 5e-6 to be effective. We use early stopping and decay learning rate to maximize the average precision on the validation set. The validation precision/recall of our filter for events with non-neutral indicators and *bei* events are given in Figure 5.

After the process of filtering lower-quality generation results, we acquire 102,218 high-precision common Chinese emotional events including 44,282 events with non-neutral indicators, 57,754 *bei* events and 182 implicit emotional events.

## 3.4 Emotional event classification

As stated earlier, indicators are necessary constituent elements of most Chinese emotional events. For emotional events with non-neutral indicators, our novel observation is that their sentiment polarity is determined by the sentiment polarity of their indicators. That is to say, if the indicator of an emotional event is positive, this emotional event is positive; otherwise, it is negative. Thus, the emotional event classification problem can be converted into an emotional event indicator classification problem which has been addressed in section 3.1. As a consequence, we can automatically classify an emotional event with a non-neutral indicator according to the classification result of the indicator it contains. The classification result of some emotional events with non-neutral indicators is reported in Table 6.

Table 6: Classification of emotional events with non-neutral indicators.

| Positive Emotional Events | Negative Emotional Events |
|---|---|
| 摆脱疾病的折磨 | 遭受痛苦 |
| (be free from the torment of illness) | (suffer pain) |
| 荣获优秀员工奖 | 面临职业发展的困境 |
| (win the Outstanding Employee Award) | (face difficulties in career development) |
| 获得家人的理解 | 错吃发霉的饼干 |
| (gain the understanding of one's family) | (accidentally eat moldy cookies) |
| 打败强大的对手 | 丢失钱包 |
| (defeat a formidable opponent) | (lose one's wallet) |
| 免于刑事处罚 | 白等一个结果 |
| (be exempt from criminal punishment) | (wait in vain for a result) |

However, for *bei* events, we are unable to carry out the event classification based on the classification result of *bei* event indicators. This is because for neutral *bei* event indicators, the emotional events containing them may have opposite polarity. For example, "被安排干脏活 (be arranged to do the dirty work)" is a negative event but "被安排做轻松的事情 (be arranged to do the easy work)" is a positive event. For the remarkable capabilities of LLMs, we also ask SparkDesk to conduct the classification task for *bei* events. We conduct a human evaluation on 100 classified positive events and negative events, respectively. The evaluation result demonstrate that SparkDesk can yield good *bei* event classification performance with a precision of 1.00, as shown in Table 7.

Table 7: Classification of *bei* events.

| **Positive *bei* Events** | **Negative *bei* Events** |
|---|---|
| 被善待 | 被拒绝 |
| (be treated with kindness) | (be refused) |
| 被体谅工作压力大 | 被禁止参加婚礼 |
| (be understood to work under high pressure) | (be forbidden to attend the wedding) |
| 被同意出国旅行 | 被没收手机 |
| (be allowed to travel abroad) | (mobile devices were confiscated) |
| 被夸奖有责任心 | 被忽悠购买减肥产品 |
| (be praised for being responsible) | (be tricked into buying weight loss products) |
| 被协助完成任务 | 被攻击头部 |
| (be assisted in completing the task) | (be hit on the head) |

Since the number of implicit emotional events is limited (182), we perform the classification task for implicit emotional events relying on human efforts. The classification result of a subset of these implicit emotional events is given in Table 8.

Table 8: Classification of implicit emotional events.

| **Positive Events** | **Negative Events** |
|---|---|
| 盈利 (make a profit) | 中毒 (be poisoned) |
| 成功 (success) | 上当 (be cheated) |
| 夺冠 (win the championship) | 失业 (lose the job) |
| 中奖 (win the lottery) | 生病 (get ill) |
| 获救 (be rescued) | 死亡 (die) |

## 4 EVALUATION

### 4.1 Intrinsic evaluation

To test the quality of the acquired emotional events in this paper, we randomly sample 10,000 emotional events (5,000 events with non-neutral indicators and 5,000 *bei* events) and conduct a human evaluation on these events. We ask three Chinese university students majoring in Chinese Language and Literature to judge whether an emotional event is valid, as they are highly familiar with Chinese emotional events. If an emotional event is linguistically and semantically correct, we assign a "valid" to it; Otherwise, we label it with "unvalid". We compute the precision as $c/(c+i)$, where c and $i$ are the counts of valid and unvalid events, respectively. The agreement between the three annotators, measured using Fleiss Kappa value [40], are 0.95 for events with non-neutral indicators and 0.92 for *bei* events. Both Fleiss Kappa scores indicate a high agreement. The evaluation results are shown in Table 9, which prove that our emotional events are of high-quality.

Table 9: Quality of emotional events.

| Events | Number of Events | Precision |
|---|---|---|
| events with non-neutral indicators | 5,000 | 0.96 |
| *bei* events | 5,000 | 0.94 |

There is no direct competitor that provides common emotional events in Chinese language. ConceptNet [11] is a commonsense knowledge graph whose nodes are concepts including events (e.g. "buy hamburger" and "get job") and edges are relations between two nodes. ConceptNet is constructed relying mostly on human intelligence and thus is known for its high quality. Also, ConceptNet is the largest publicly available commonsense knowledge graph. Therefore, we compare the emotional event coverage of our work with ConceptNet. Specifically, we adopt the Chinese subset of ConceptNet 5.5 which involves 260,662 nodes including head nodes and tail nodes, for comparison. The comparison result reported in Table 10 shows that ConceptNet has an extremely low emotional event coverage of 0.01, which validates that existing commonsense knowledge graphs are limited to emotional events and our acquired common emotional events can be used to enrich existing commonsense knowledge graphs by adding novel emotional event nodes.

Table 10: Emotional event coverage comparison with ConceptNet.

| Knowledge Bases | Total Number of Nodes | Number of Emotional Event Nodes | Emotional Event Coverage |
|---|---|---|---|
| ConceptNet (Chinese) | 260,662 | 2,571 | 1% |
| Ours | 102,036 | 102,036 | 100% |

### 4.2 Extrinsic evaluation

As stated before, ECE is a subtask of emotion analysis [41], which aims at detecting potential causes given an emotion. It is a much more difficult task compared to emotion detection or emotion classification. Researchers have proposed numerous models for this task, with PAE-DGL [42], RTHN [30], RHNN [29], CANN [27], MANN [28], EF-BHA [43], CNCM [8] and UECA-Prompt [44] being eight representative models for emotion cause identification, and their codes are publicly available. However, all these models neglect emotional events. In order to prove the usefulness of the emotional events we acquire in this paper, we augment these ECE models with them. The augmented models are denoted as EPAE-DGL, ERTHN, ERHNN, ECANN, EMANN, EEF-BHA, ECNCM and EUECA-Prompt, respectively. We measure these models and their augmented models on a Chinese emotion cause corpus.

### 4.2.1 Dataset.

[45] explores the causes of explicit emotions by constructing a Chinese emotion cause corpus, which contains 10,603 instances from news texts. Each instance has only one emotion keyword and one or more emotion causes. For the emotion in an instance, the causes are annotated manually. "1999年春天，平邑县白彦镇大营村支部书记王某因与村主任桂某关系不和，<cause>遭到桂某等人殴打</cause>，王某的儿子心生<keyword>怨恨</keyword>，又伙同他人将桂某殴打致轻伤，随后，桂向法院提起自诉，但因事实不清、证据不足而撤诉 (In the spring of 1999, Wang, the party branch secretary of Daying Village in Baiyan Town, Pingyi County, had a poor relationship with the village head, Gui. As a result, he <cause>was beaten by Gui and others</cause>. Wang's son felt <keyword>disgusted</keyword>, together with others, beat Gui, causing him minor injuries. Later, Gui filed a self-complaint with the court, but the case was dismissed due to unclear facts and insufficient evidence)" is an instance from the dataset. We parse each sentence in an instance using jieba [46], a popular Chinese word segmentation tool and part-of-speech (POS) tagger.

### 4.2.2 Implementation details.

For PAE-DGL, RTHN, RHNN, CANN, MANN, EF-BHA, CNCM and UECA-Prompt, the parameter settings are consistent with [42], [30], [29], [27], [28], [43], [8] and [44], respectively. These eight models and their corresponding augmented models share the same parameter settings. With regard to emotional event infusion, we consider whether a clause contains at least one emotional event indicator as a new dimension or a new feature. If a clause involves one or more indicators, the feature value is set to 1; otherwise, it is set to 0. Regarding PAE-DGL, RTHN, RHNN, CANN and MANN, the event integration is carried out by appending the new dimension to the clause vector. Specifically, given an input clause $S = [w_0, w_1, w_2, ..., w_n]$ and the feature value 1 or 0, we take $S' = [w_0, w_1, ..., w_n, 1]$ or $S' = [w_0, w_1, ..., w_n, 0]$ as the final clause embedding.

In EPAE-DGL, we concatenate the final clause vector and the Dynamic Global Label vector as the feature for emotion cause prediction. The concatenation of the final clause embedding and positional embedding is fed to the multi-head self-attention layer of ERTHN. In ERHNN, ECANN and EMANN, the final clause embedding is entered into the clause attention layer, co-attention layer and multi-attention mechanism layer, respectively. For EEF-BHA, a 768-dimensional vector is initialized at random for the new feature, which is then integrated with the clause vector. In other words, the final embedding representation of the clause is the sum of the clause vector and the feature vector. The final clause vector is entered into the word-level context attention module. Regarding ECNCM and EUECA-Prompt, the new feature is incorporated by first randomly initializing a vector for it and then integrating the initialized vector with the clause vector. The initialized vector is 64-dimensional in ECNCM and 768-dimensional in EUECA-Prompt. We feed the final clause vector into the Bi-LSTM layer of ECNCM and the mask prediction layer of Bert in EUECA-Prompt. Following [39], we randomly split the data with a 9:1 ratio, using 9 folds as training data and the remaining 1 fold as testing data. The reported results are the average values from 10-fold cross-validation.

### 4.2.3 Evaluation metrics.

We evaluate the performance of a ECE model by precision, recall and F-score, which are commonly used evaluation metrics in emotion cause detection [4][31]. The precision, recall and F-score are defined as follows.

$$\text{Precision} = \frac{\Sigma \text{correct causes}}{\Sigma \text{proposed causes}} \qquad (1)$$

$$\text{Recall} = \frac{\Sigma \text{correct causes}}{\Sigma \text{annotated causes}} \qquad (2)$$

$$\text{F-score} = \frac{2 \times \text{Precision} \times \text{Recall}}{\text{Precision} + \text{Recall}} \qquad (3)$$

Where $\Sigma$correct causes is the number of correctly predicted causes, $\Sigma$proposed causes represents the number of predicted causes and $\Sigma$annotated causes denotes the amount of causes annotated in all the 10,603 instances.

### 4.2.4 Baseline models.

We integrate our emotional events into eight ECE models: 1) PAE-DGL: A reordered model that incorporates relative position information and dynamic global label for the emotion cause identification task [42]; 2) RTHN: A joint ECE framework, which is composed of a lower word-level encoder based on RNNs to encode multiple words in each clause and an upper clause-level encoder based on Transformer to learn the correlation between multiple clauses in a document [30]; 3) RHNN: A regularized hierarchical neural network for emotion cause analysis, which combines the discourse context information and knowledge-based regularizations [29]; 4) CANN: A deep learning method that exploits attention mechanism to enhance text representations by modeling the mutual impacts between an emotion clause and the other clauses for the emotion cause recognition task [27]; 5) MANN: A multi-attention-based neural network that realizes the emotion cause attention and candidate clause attention for ECE [28]; 6) EF-BHA: A method for ECE task by capturing the document-level context and filtering irrelevant information [43]; 7) CNCM: The baseline method that extracts emotion causes exploiting causal narrative information [8]; 8) UECA-Prompt: A deep learning method that attempts to solve different emotion cause prediction tasks with prompt. Moreover, this method designs a directional constraint module and a sequential learning module to mitigate the dataset bias caused by position information [44].

Table 11: Performance of the baseline systems and their emotional event-enhanced systems.

| Method | Precision | Recall | F-score |
|---|---|---|---|
| PAE-DGL | 75.25 | 59.31 | 66.23 |
| **EPAE-DGL** | **78.35** | **59.65** | **67.52** |
| RTHN | 73.51 | 68.10 | 70.51 |
| **ERTHN** | **74.61** | **69.71** | **72.00** |
| RHNN | 77.42 | 68.92 | 72.92 |
| **ERHNN** | **78.83** | **69.42** | **73.83** |
| CANN | 76.43 | 59.22 | 66.73 |
| **ECANN** | **77.21** | **60.74** | **67.99** |
| MANN | 74.64 | 67.22 | 70.74 |
| **EMANN** | **75.76** | **68.63** | **72.02** |
| EF-BHA | 75.15 | 69.84 | 72.40 |
| **EEF-BHA** | **75.62** | **70.45** | **72.94** |
| CNCM | 84.66 | 70.84 | 77.14 |
| **ECNCM** | **85.87** | **71.62** | **78.10** |
| UECA-Prompt | 74.88 | 74.73 | 74.64 |
| **EUECA-Prompt** | **77.69** | **74.99** | **75.89** |

### 4.2.5 Experimental results.

We evaluate the ECE performance of these baseline models before and after common emotional event infusion. The results are reported in Table 11. We observe a common trend of increase in across all evaluation metrics when common emotional events are injected into these models. This demonstrates the important role that common emotional events play in identifying emotion causes. EPAE-DGL significantly outperforms PAE-DGL in terms of precision and the improvement is more than 3%. ERTHN reports an obvious improvement over all metrics, with a 1.1%, 1.6% and 1.5% increase in precision, recall and F-score, respectively. Similar to ERTHN, EMANN exhibits a 1.1% increase in precision, 1.6% increase in recall and 1.2% increase in F-score. ERHNN shows a 1.4% increase in precision. However, there is still room for improvement in recall. ECANN outperforms CANN with a 1.5% increase in recall and 1.2% increase in F-score, while the precision improvement is modest. Though the improvement is not significant, EEF-BHA still achieves a higher precision, recall and F-score than EF-BHA. ECNCM achieves a much higher Precision score (85.87%) than other methods, but without reducing the Recall score. For UECA-Prompt, emotional event injection results in a remarkable gain of 2.8% in precision. We argue that the emotional events injected into these models contribute the improvements. As stated earlier, emotional events are important causes of human emotions, for the emotional event-enhanced models can capture more emotional information and thus to extract emotion causes more correctly.

## 5 CONCLUSION

This paper proposes a novel methodology for acquiring common Chinese emotional events. We first collect a list of emotional event indicators which are linguistic features of Chinese emotional events. With these indicators, we attempt to acquire emotional events by prompting a large language model to generate phrases that satisfy a set of constraints. Also, we categorize these obtained emotional events into two groups (positive events or negative events) exploiting different methods. The resulting large-scale commonsense knowledge base consists of more than 100,000 emotional events. To the best of our knowledge, this is the first resource of enormous common Chinese emotional events with affective polarity labels. Intrinsic evaluation results show that the emotional events we acquire in this paper are of high-quality. The experiments performed on a Chinese emotion cause dataset also demonstrates that the obtained common emotional events are highly beneficial to the field of emotion cause detection. We argue that the common emotional events in this paper provide the foundation for future research on sentiment analysis, such as the identification of implicit emotions or emotion causes.

One limitation of our work is that the indicator list we construct is not complete, as acquiring Chinese emotional event indicators especially *bei* event indicators is a labor-intense and time-consuming task. We will enrich our indicator list in future work.